\documentclass[letterpaper, 10 pt, conference]{ieeeconf}
\IEEEoverridecommandlockouts                              % This command is only needed if you want to use the \thanks command

\usepackage{pdfpages}
\usepackage{amsmath}
\usepackage{graphics} % for pdf, bitmapped graphics files
\usepackage{epsfig} % for postscript graphics files
\usepackage{mathptmx} % assumes new font selection scheme installed
\usepackage{times} % assumes new font selection scheme installed
\usepackage{amsmath} % assumes amsmath package installed
\usepackage{dirtytalk}
\usepackage{amssymb}  % assumes amsmath package installed
\usepackage{color}
\usepackage{caption}
\usepackage{subcaption}
\usepackage{diagbox}
\usepackage{algorithm}
\usepackage{algpseudocode}
\usepackage{makecell}
\usepackage{array}
\usepackage{svg}

\usepackage{booktabs}
\title{\LARGE \bf
GenDDS: Generating Diverse Driving Video Scenarios with Prompt-to-Video Generative Model}

\author{
Yongjie Fu, Yunlong Li, and Xuan Di$^{*} 
 ~\IEEEmembership{Member,~IEEE}$% <-this % stops a space
\thanks{$^{*}$\textit{Corresponding author: Xuan Di.}
}
\thanks{$^{\text{‡}}$This work is sponsored by NSF CPS-2038984 and NSF ERC-2133516.}
\thanks{Yongjie Fu is with the Department of Civil Engineering and Engineering Mechanics,
        Columbia University, New York, NY, 10027, USA (E-mail: yf2578@columbia.edu).}
\thanks{Yunlong Li is with the Department of Electrical Engineering, Columbia University, New York, NY, 10027, USA (E-mail: yl5330@columbia.edu).}
\thanks{Xuan Di is with the Department of Civil Engineering and Engineering Mechanics, Columbia University, New York, NY, 10027 USA, and also with the Data Science Institute, Columbia University, New York, NY, 10027 USA (E-MAIL: sharon.di@columbia.edu).}
\thanks{$**$ This work has been accepted by 27th IEEE International Conference on Intelligent Transportation Systems (ITSC 2024).}
}

\begin{document}

\maketitle
\thispagestyle{empty}
\pagestyle{empty}

%%%%%%%%%%%%%%%%%%%%%%%%%%%%%%%%%%%%%%%%%%%%%%%%%%%%%%%%%%%%%%%%%%%%%%%%%%%%%%%%
\begin{abstract}

Autonomous driving training requires a diverse range of datasets encompassing various traffic conditions, weather scenarios, and road types. Traditional data augmentation methods often struggle to generate datasets that represent rare occurrences. To address this challenge, we propose GenDDS, a novel approach for generating driving scenarios generation by leveraging the capabilities of Stable Diffusion XL (SDXL), an advanced latent diffusion model. Our methodology involves the use of descriptive prompts to guide the synthesis process, aimed at producing realistic and diverse driving scenarios. With the power of the latest computer vision techniques, such as ControlNet and Hotshot-XL, we have built a complete pipeline for video generation together with SDXL. We employ the KITTI dataset, which includes real-world driving videos, to train the model. Through a series of experiments, we demonstrate that our model can generate high-quality driving videos that closely replicate the complexity and variability of real-world driving scenarios. This research contributes to the development of sophisticated training data for autonomous driving systems and opens new avenues for creating virtual environments for simulation and validation purposes. 

\end{abstract}

%%%%%%%%%%%%%%%%%%%%%%%%%%%%%%%%%%%%%%%%%%%%%%%%%%%%%%%%%%%%%%%%%%%%%%%%%%%%%%%%
\section{INTRODUCTION}
\label{sec-intro}
In the rapidly evolving landscape of artificial intelligence, generative models have emerged as a cornerstone technology, significantly impacting various industries, including autonomous driving and smart city \cite{ruan2022learning, fu2023federated}. These advanced models, capable of synthesizing highly realistic and diverse data, are revolutionizing the way we approach the development and testing of autonomous vehicles. By generating intricate driving scenarios, they offer a scalable solution to the challenge of training and validating autonomous systems under a wide array of conditions. This capacity not only accelerates the advancement of autonomous driving technologies but also enhances their safety and reliability, marking a pivotal shift in our journey toward fully autonomous transportation systems.

The past year has witnessed significant advancements in deep generative models, spanning multiple data areas, including natural language, audio, and visual content. The domain of self-driving technology increasingly requires training data of superior quality that is well-annotated.  The methods for obtaining autonomous driving datasets often rely on simulators or image generation models to expand the dataset. However, capturing datasets under specific environmental conditions or unique scenarios can be challenging using these methods. Data generated by simulators is not as realistic compared to the real-world data. As a result, to create a more straightforward and realistic generator for autonomous driving datasets, we propose GenDDS, which utilizes diffusion and text-to-video models for this task.

\subsection{Related Work}
Video generation models utilize neural networks to generate video samples. GAN-based (generative adversarial network \cite{goodfellow2014generative, mo2022quantifying}), autoregressive-based, VAE-based (variational autoencoder \cite{kingma2013auto}), normalizing flow \cite{mo2022trafficflowgan}, and diffusion-based generators are widely used in data generation. Diffusion probabilistic models (DM) have emerged as a state-of-the-art technique across several domains, notably in image and video generation applications. For images, diffusion models have been successfully applied to tasks such as generation, editing, and image-to-image translation \cite{li2022vqbb}. In the realm of video, recent advancements in diffusion models have significantly improved video quality and realism. Applications include general video generation \cite{harvey2022flexible}, as well as the generation of video from text prompts \cite{villegas2022phenaki}, demonstrating enhanced capabilities in creating dynamic and realistic video sequences.

Text-to-image models are a popular kind of controllable generation model, that can generate high-resolution, multi-styled images. Various strategies have been explored to extend the quality and efficiency of generation. The transformer architecture is widely utilized across various tasks \cite{mao2022trace, fu2024digital}. Low-Rank Adaptation (LoRA) \cite{hu2021lora} for the transformer architecture is a useful fine-tuning strategy for the diffusion model.

Existing research has employed diffusion-based models within the realms of autonomous driving and transportation. Specifically, the guided conditional diffusion model has been leveraged to create controllable traffic simulations, allowing users to dictate the desired properties of trajectories  \cite{zhong2023guided}. Li introduces DrivingDiffusion \cite{li2023drivingdiffusion} as a technique for generating multi-view videos of autonomous driving scenes, featuring precise layout control. Harvey \cite{harvey2022flexible} introduces a framework based on Diffusion Probabilistic Models (DDPMs) that is capable of generating extended video sequences with realistic and coherent scene completion.

In this study, we propose GenDDS to generate contextually relevant driving videos based on textual descriptions. Our methodology leverages the KITTI driving dataset, a comprehensive collection of real-world automotive driving scenarios, to train a LoRA (Low-Rank Adaptation) layer on top of the pre-existing SDXL model. This adaptation allows for the efficient incorporation of dynamic, text-based requirements into the video generation process.

Following the initial training phase, the enhanced SDXL model, equipped with the trained LoRA layer, is employed within the Hotshot XL framework for the generation of images. To enrich the model's capability in handling diverse driving scenarios and improve its generalization, we introduce an additional component known as ControlNet. ControlNet utilizes auxiliary driving videos, beyond those found in the KITTI dataset, to provide contextual grounding and control signals during the image synthesis process.

The highlight of GenDDS is a sophisticated video generation model capable of producing high-quality driving videos that are not only visually compelling but also accurately reflect the nuances of the provided textual descriptions. This approach demonstrates a significant advancement in the field of conditional video synthesis, particularly in the domain of autonomous driving simulation and training environments.

\subsection{Contributions of this work}

We propose GenDDS, a driving video generation method. The main contributions of the  framework can be summarized as follows: 
\begin{enumerate}
    \item Utilize the generative model API to create text prompts from video inputs together with manual creation, streamlining the process of preparing training data. 
    \item Fine-tune the LoRA-based Stable Diffusion XL (SDXL) model on a real-world driving dataset.
    \item Load the SDXL model in Hotshot-XL and generate driving scenarios given different text prompt conditions together with ControlNet.
   
\end{enumerate}
The rest of the paper is organized as follows. Sec.~\ref{sec-preliminory} introduces the preliminary knowledge used in this paper. Sec.~\ref{Solution-approach} illustrates the solution approach. Sec.~\ref{sec:experiments} introduces the setting and results of the experiments. And Sec~.\ref{sec:conclusion} concludes this study.

\section{Preliminory}
\label{sec-preliminory}
\subsection{Stable Diffusion}
Diffusion Models (LDM) \cite{sohl2015deep} represent an approach in the field of generative modeling, particularly in the generation of a data distribution $p(x)$, by operating within a latent space rather than the original data space. The core concept behind LDMs lies in their ability to learn a distribution over a compressed, or latent, representation of data, which significantly reduces computational costs and improves model efficiency. A key equation that underlies the LDM framework is the diffusion process, which is modeled as a Markov chain of Length $T$ that gradually adds noise to the data over a sequence of steps until it reaches a Gaussian distribution. LDM can be interpreted as an equally weighted sequence of demonizing autoencoders  $\boldsymbol{e}_{\theta}(x_t, t); t = 1 \ldots T$, which are trained to predict a denoised variant of their input $x_t$, where $x_t$ is a noisy version of the input $x$. The corresponding objective can be simplified to:

\begin{equation}
    L_{DM} = \mathbb{E}_{x_t,\epsilon \sim \mathcal{N}(0,1),t}\left[ \left\lVert \epsilon - \boldsymbol{e}_{\theta}(x_t, t) \right\rVert^2 \right]
\end{equation}

with $t$ uniformly sampled from $\{1, \ldots, T\}.$\\.

% \begin{figure}[H]
% 	\centering
% 	\includegraphics[scale=.2]{figures/simple-structure.jpg}
% 	\centering 
% 	\caption{Simple Diffusion Architecture with text prompt as input (\tcb{redraw})}
% 	\label{fig:stable-diffusion}
% \end{figure}
The Stable Diffusion model is a more flexible, conditional image generator that utilizes a U-Net backbone and cross-attention mechanisms. It is effective for learning models that leverage attention across various input modalities. Diffusion models are able to model conditional distributions of the form $p(z|y)$, which can be implemented with a conditional demonizing autoencoder to control the synthesis process through condition $y$ such as text prompts.

Diffusion models are enhanced to become more versatile generators of conditional images by augmenting their foundational UNet structure, which is known for its efficacy with the cross-attention mechanism cited in the work. This mechanism is adept at learning from models that base attention on a range of input types, as indicated in references. In order to process the input $y$ from diverse sources, such as verbal instructions, a domain-specific encoder $\tau_\theta$ is utilized to map $y$ into a preliminary representation  $\tau_\theta(y) \in \mathbb{R}^{M \times d_r}$. This is subsequently applied to the intermediate layers of the UNet by employing a cross-attention layer. The cross-attention layer is defined as:

\begin{equation}
    \text{Attention}(Q, K, V) = \text{softmax}\left(\frac{QK^T}{\sqrt{d}}\right) \cdot V
\end{equation}

Where the query $Q$, key $K$, and value $V$ matrices are produced as follows:

\begin{align*}
    Q &= W^Q_i \cdot \varphi_i(z_t), \\
    K &= W^K_i \cdot \tau_\theta(y), \\
    V &= W^V_i \cdot \tau_\theta(y).
\end{align*}
Where $\varphi_i$ denotes a representation of UNet implementing and $W^Q_i, W^K_i, W^V_i$ are learnable projection matrices.
\begin{figure}[H]
	\centering
	\includegraphics[scale=.35]{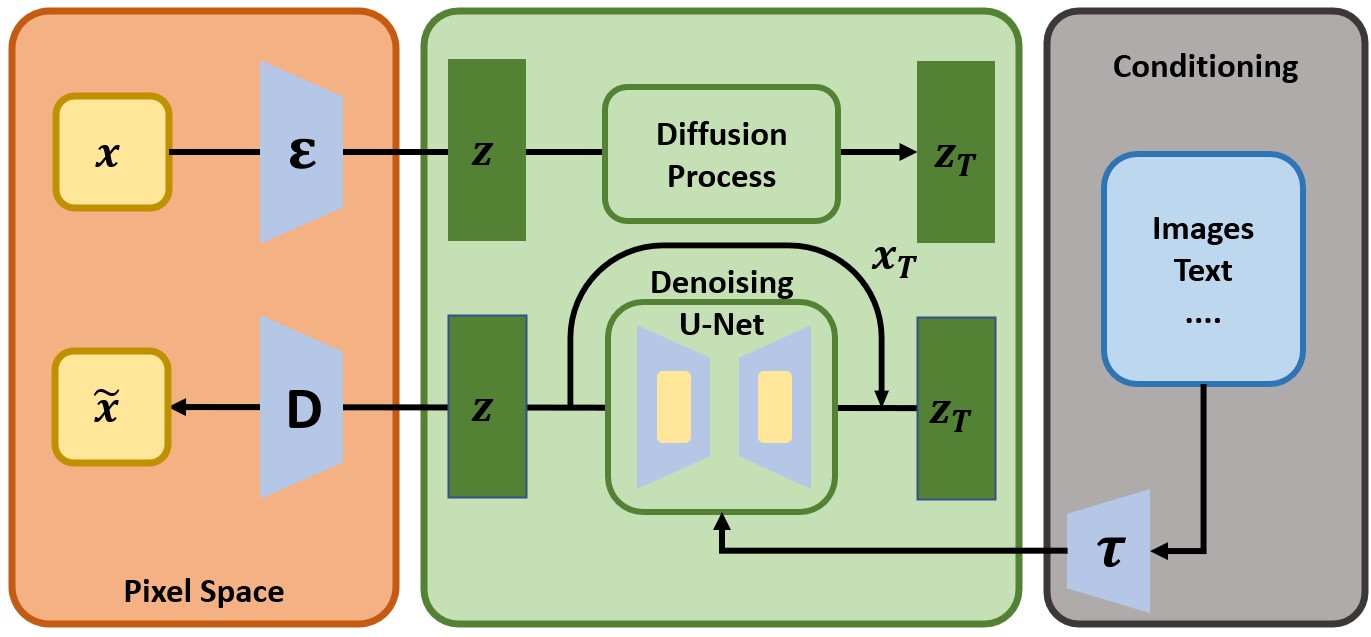}
	\centering 
	\caption{Stable Diffusion Architecture}
	\label{fig:stable-diffusion}
\end{figure}

\subsection{Stable Diffusion XL (SDXL)}

Stable Diffusion XL \cite{podell2023sdxl} is a latent diffusion model designed for the task of generating images conditioned on text. The distinction of SDXL lies in its three times larger U-Net backbone, which provides more attention blocks and a larger cross-attention context. SDXL shows improved performance compared to the stable diffusion model.

SDXL utilizes a heterogeneous distribution of transformer blocks within the U-Net. Additionally, it opts for a more powerful pre-trained text encoder (OpenCLIP ViT-BigG in combination with CLIP ViT-L \cite{ilharco_gabriel_2021_5143773}) for text conditioning. This modification results in a U-Net model size of 2.6 billion parameters, compared to the stable diffusion model's 865 million parameters.
\subsection{Low-Rank Adaptation (LoRA)}
LoRA \cite{hu2021lora} locks the weights of the pre-trained model and introduces trainable matrices based on rank decomposition into every layer of the transformer structure. This technique significantly lowers the count of parameters that need to be trained for subsequent tasks, which can work with any diffusion model.

The updates of the weights have a low ``intrinsic rank \cite{aghajanyan2020intrinsic}'' during adaptation. For an initially established weight matrix $W_0 \in \mathbb{R}^{d \times k}$, its modification is limited through a low-rank factorization, expressed as $W_0 + \Delta W = W_0 + BA$. Here, $B \in \mathbb{R}^{d \times r}$ and $A \in \mathbb{R}^{r \times k}$, where the rank $r$ is significantly smaller than the smaller dimension between $d$ and $k$. In the training process, $W_0$ remains static and does not undergo gradient updates, while $A$ and $B$ are adjustable parameters. It's important to note that both $W_0$ and $\Delta W = BA$ interact with the identical input, and their output vectors are added together element-wise. The updated forward pass formula is then given by $h = W_0x$, indicating:

\begin{equation}
h = W_0x + \Delta Wx = W_0x + BAx
\end{equation}

\subsection{ControlNet}

ControlNet is an innovative neural network architecture designed to inject additional conditional controls into diffusion models. ControlNet will inject additional conditions into the neural network, which can give control to a large pre-trained diffusion model. The ControlNet is applied to each of the encoder levels of the U-Net. A trainable copy of the encoding blocks and the middle block of the diffusion model is created. The ControlNet is computationally efficient since the locked copy parameters are frozen, no gradient computation is required in the originally locked encoder for the fine-tuning. ControlNet is computationally efficient because its locked copy parameters are frozen, eliminating the need for gradient computation in the originally locked encoder during fine-tuning.

\subsection{Hotshot-XL}

\begin{figure}[H]
	\centering
	\includegraphics[scale=.57]{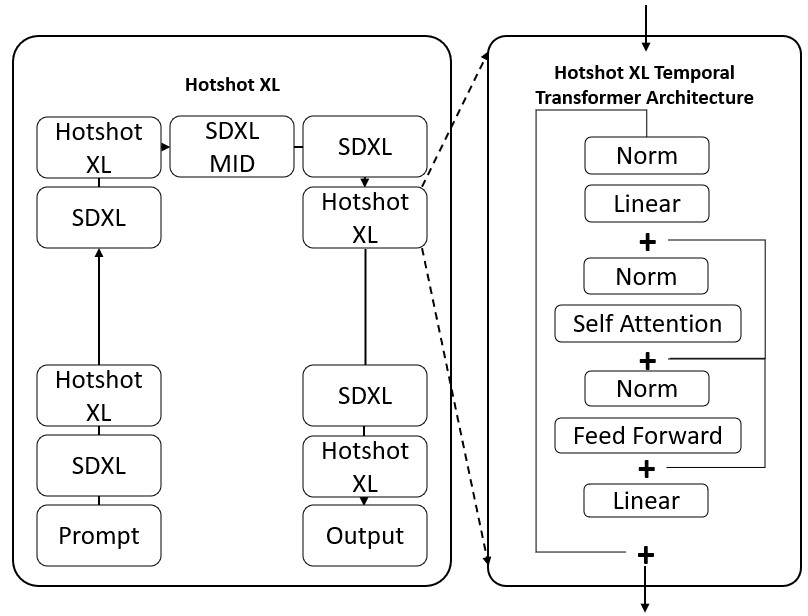}
	\centering 
	\caption{Hotshot-XL Architecture}
	\label{fig:hotshot}
\end{figure}

Hotshot-XL \cite{Mullan_Hotshot-XL_2023} is a text-to-GIF generation model that can work together with SDXL. We are able to feed any fine-tuned SDXL model into Hotshot-XL. Loading the fine-tuned SDXL-based LoRAs is easier than fine-tuning Hotshot-XL. The architecture of Hotshot-XL is illustrated in Figure~\ref{fig:hotshot}. Hotshot-XL consists of multiple SDXL modules and a Hotshot-XL temporal transformer module, which are connected sequentially to take a prompt as input and output the GIFs. The temporal transformer architecture is detailed on the right side of Figure~\ref{fig:hotshot}, consisting of normalization, self-attention, and feed-forward neural networks. The Hoishot-XL can generate high-resolution GIFs at 8 frames per second and supports various image sizes.

\section{Proposed Pipline}
\label{Solution-approach}

We aim to address the challenge of generating realistic driving videos straightforwardly using a generative model. Consequently, we propose a pipeline that leverages the capabilities of SDXL to generate a driving video dataset.

\begin{figure*}[htb]
	\centering
	\includegraphics[scale=.41]{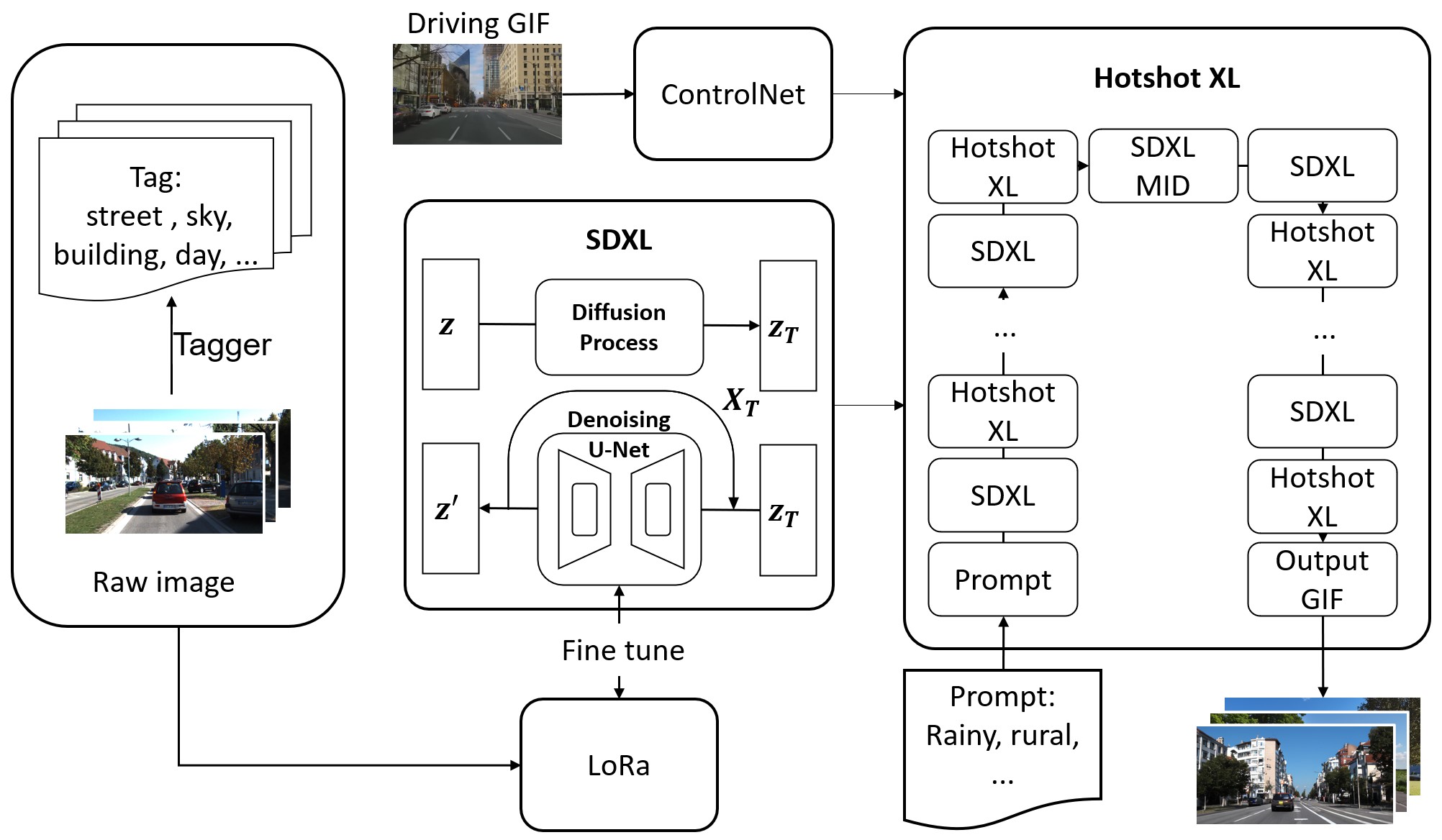}
	\centering 
	\caption{Pipeline for the training process of GenDDS.}
	\label{fig:pipeline}
\end{figure*}

Figure.~\ref{fig:pipeline} illustrates how GenDDS integrates the SDXL, LoRA, ControlNet auto-tagger process, and Hotshot-XL. We extract raw images from open datasets and use Tagger to generate tags for each frame. These prepared datasets are then used to train the LoRA layers in SDXL. Once the fine-tuned SDXL model is ready, we feed it into Hotshot-XL, which acts as a module in the sequence. Concurrently, we use ControlNet to enhance the spatial relationships generated by Hotshot-XL. After completing the entire pipeline, we can run inference with Hotshot-XL to generate driving videos. We will illustrate the details of each module in the experiment section.

\section{Experiments}
\label{sec:experiments}

\subsection{Datasets}

The main dataset used in our experiments is the KITTI dataset \cite{Geiger2012CVPR}, a set of benchmarks for computer vision algorithms mainly used in autonomous driving scenarios. The KITTI dataset is created by the Karlsruhe Institute of Technology and the Toyota Technological Institute in Chicago. Ranging from stereo images to optical flow, from visual odometry to 3D object detection and tracking. In our project, we utilize the high-resolution images and sequences from this dataset to train our model to generate realistic driving scenarios under various environmental conditions.

\subsection{Experiment Setup}
Our experiments utilize the functionality of the SDXL base model \cite{podell2023sdxl}, which is known for its ability to generate high-resolution images through deep learning techniques. We enhanced this model by integrating Low-Rank Adaptation (LoRA), a modification designed for efficient parameter updating. This adaptive technique facilitates specialized training of the large-scale model for specific tasks, thus greatly reducing the need for full retraining.

A component of the data preparation was the use of the WD 1.4 MOAT Tagger V2 \cite{yang2023moat} tagging mechanism. This tagging system works by labeling each frame with the data that the image contains, such as ``outdoors, road, scenery, sky...". The automated tagging is followed by a careful manual review to ensure the accuracy of the tagging and the distinction between different images, including traffic conditions and other keywords such as ``light traffic" or ``heavy traffic". This labeling strategy improves model training efficiency by providing highly accurate annotations. These tags help train the model to recognize and generate nuances in driving scenarios based on different environments and traffic conditions.

The computational backbone of our training infrastructure is equipped with NVIDIA 3090Ti GPUs. Our training process employs a selected set of hyperparameters to optimize the balance between model accuracy and computational resource utilization. Model optimization was performed with the AdamW8bit optimizer, employing a fine-grained learning rate strategy starting at 0.0001. This approach is complemented by applying different learning rates to the U-Net and text encoder portions of the model, thus enabling customized learning trajectories within the model architecture. 

One aspect of our model configuration is the selection of training parameters to improve efficiency and effectiveness. We set the model to run 32 iteration steps per image, a decision aimed at achieving a balance between image quality and computational demands. We used the DPM++ 2M Karras sampler and set the CFG ratio to 7.5, which improves the model's ability to generate images that are consistent with the textual descriptions and input conditions, thus maintaining the thematic integrity of the generated scenes.

\subsection{Experiment Process}
In our experimental framework, we employ real-world driving footage as ControlNet \cite{zhang2023adding}. This innovative approach utilizes the realism of actual road conditions captured on video to guide the image synthesis process. By integrating SDXL models enhanced with LoRA technology into the Hotshot XL environment \cite{Mullan_Hotshot-XL_2023}, we set the stage for generating video content. This approach allowed us to generate continuous frame sequences that effectively simulate dynamic driving scenarios.

In ControlNet, depth information is estimated from real traveling video inputs by depth detection methods. This depth information plays a role throughout the image generation process, guiding the model to maintain accurate spatial relationships and perspectives in the generated image. The integration of depth estimation ensures that the synthetic enhancements applied during the generation process are not only visually realistic, but also contextually appropriate and reflective of real driving scenarios.

The core of this process revolved around using the real driving videos as a baseline from which the model could generate enhanced or varied scenarios under controlled conditions. The Hotshot XL facilitated this by providing the architecture to process the SDXL-LoRA model efficiently. Through this setup, we could manipulate the original video frames, applying synthetic alterations that represent different weather conditions, traffic densities, and other environmental variables, all while maintaining the original video's structural integrity.

\subsection{Results}

The experiment results for GenDDS are visualized in a series of graphs that illustrate the model's ability to synthesize highly realistic, contextually coherent video sequences under a variety of conditions. From contextual coherence to environmental adaptability, each diagram emphasizes a different aspect of the model's capabilities.

\begin{figure}[ht]
\centering
\begin{minipage}{.45\columnwidth}
\centering
\includegraphics[width=4cm]{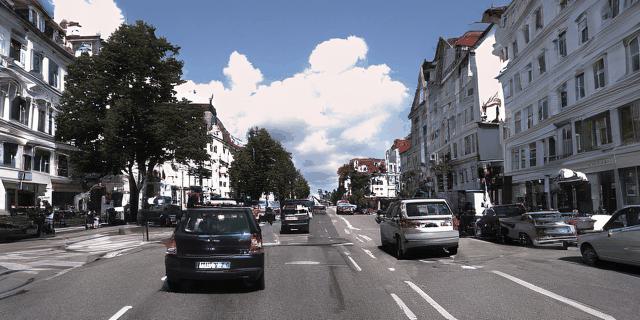}
\caption*{Frame 1}
\end{minipage}
\hfill
\begin{minipage}{.45\columnwidth}
\centering
\includegraphics[width=4cm]{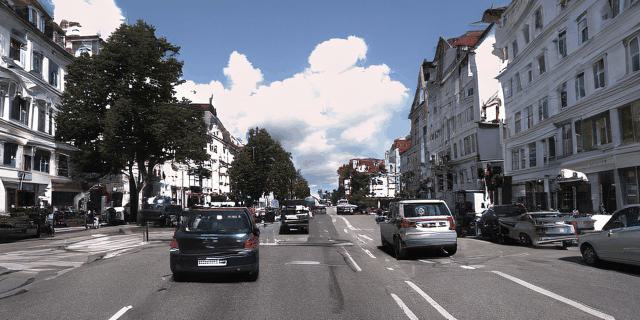}
\caption*{Frame 2}
\end{minipage}

\vspace{3mm} 

\begin{minipage}{.45\columnwidth}
\centering
\includegraphics[width=4cm]{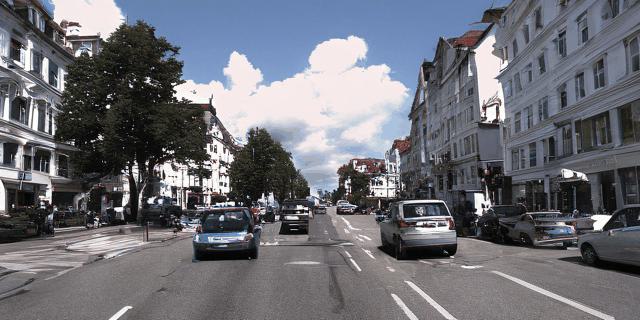}
\caption*{Frame 3}
\end{minipage}
\hfill
\begin{minipage}{.45\columnwidth}
\centering
\includegraphics[width=4cm]{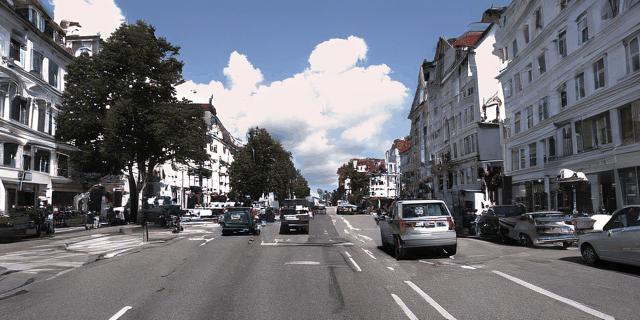}
\caption*{Frame 4}
\end{minipage}
\caption{Contextual Consistency in Sequential Frames}
\label{fig:sunny_day_frames}
\end{figure}

Figure. \ref{fig:sunny_day_frames} provides a snapshot of our model's capability to maintain contextual coherence across sequential frames. It showcases a series of four continuous frames where the model has successfully captured the dynamic nature of the driving scenario. Notably, the forward progression of vehicles in the scene is realistically rendered, with a particular focus on a vehicle that gradually moves forward across the frames. This illustrates the model's adeptness at understanding and applying the temporal continuity essential for video generation, ensuring that each frame is a logical progression from the last.

\begin{figure}[ht]
\centering
\begin{minipage}{.45\columnwidth}
\centering
\includegraphics[width=4cm]{figures/output/sunny_frames_0.jpg}
\caption*{Sunny}
\end{minipage}
\hfill
\begin{minipage}{.45\columnwidth}
\centering
\includegraphics[width=4cm]{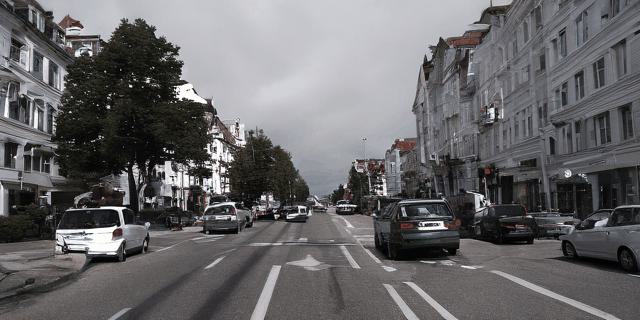}
\caption*{Cloudy}
\end{minipage}

\vspace{3mm} 

\begin{minipage}{.45\columnwidth}
\centering
\includegraphics[width=4cm]{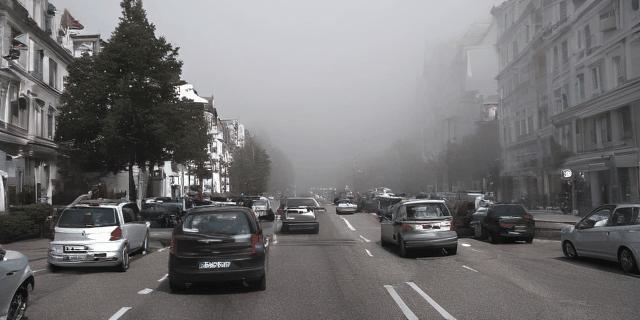}
\caption*{Foggy}
\end{minipage}
\hfill
\begin{minipage}{.45\columnwidth}
\centering
\includegraphics[width=4cm]{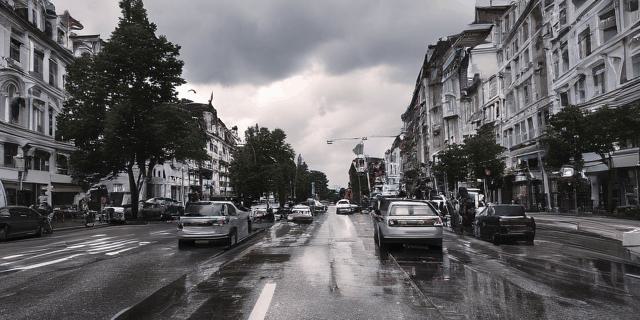}
\caption*{Rainy}
\end{minipage}
\caption{Adaptability to Different Weather Conditions}
\label{fig:weather_scenarios}
\end{figure}

In Figure.~\ref{fig:weather_scenarios}, our model is able to generate video sequences under different weather conditions, including sunny, cloudy, foggy, and rainy days, thus highlighting the versatility of the model. This variety of outputs demonstrates the model's broad understanding of environmental factors and its ability to modify the scene accordingly. From changing lighting conditions to introducing elements such as rain or fog, each weather condition presents a set of challenges that our model handles with ease, giving us a glimpse of its great adaptability.

\begin{figure}[ht]
\centering
\begin{minipage}{.45\columnwidth}
\centering
\includegraphics[width=4cm]{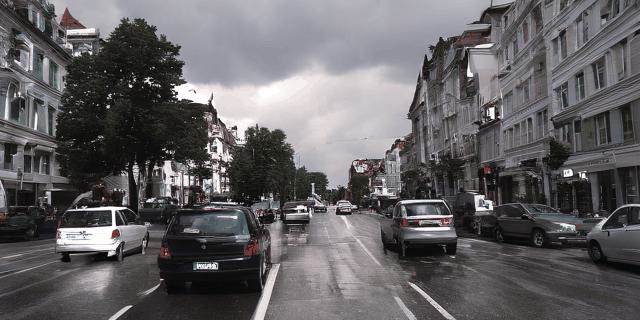}
\caption*{Very heavy}
\end{minipage}
\hfill
\begin{minipage}{.45\columnwidth}
\centering
\includegraphics[width=4cm]{figures/output/rain_car-_frames_0.jpg}
\caption*{Heavy}
\end{minipage}

\vspace{3mm}

\begin{minipage}{.45\columnwidth}
\centering
\includegraphics[width=4cm]{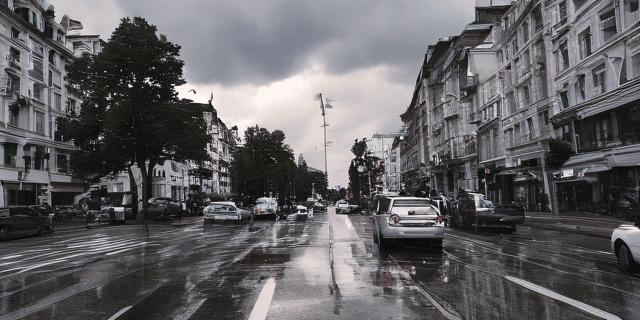}
\caption*{Moderate}
\end{minipage}
\hfill
\begin{minipage}{.45\columnwidth}
\centering
\includegraphics[width=4cm]{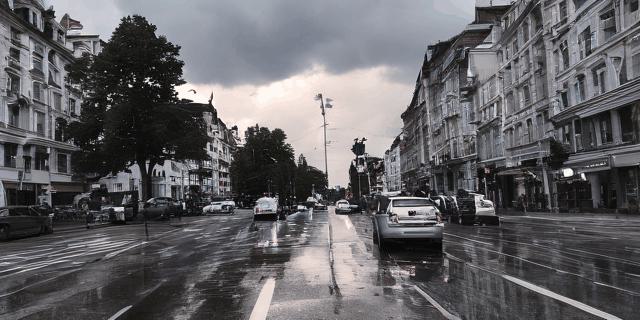}
\caption*{Light}
\end{minipage}
\caption{Generating Variability in Traffic Density}
\label{fig:traffic_scenarios}
\end{figure}

Building upon the rain scenario, Figure.~\ref{fig:traffic_scenarios} explores the model's ability to generate videos with different traffic densities. This feature is particularly noteworthy as it demonstrates the model's capacity for fine-grained control over the elements within the scene. By adjusting the number of vehicles, our model can simulate a range of traffic conditions from light to heavy, offering insights into its potential for creating varied driving scenarios that are crucial for training autonomous driving systems.

\begin{figure}[ht]
\centering
\begin{minipage}{.45\columnwidth}
\centering
\includegraphics[width=4cm]{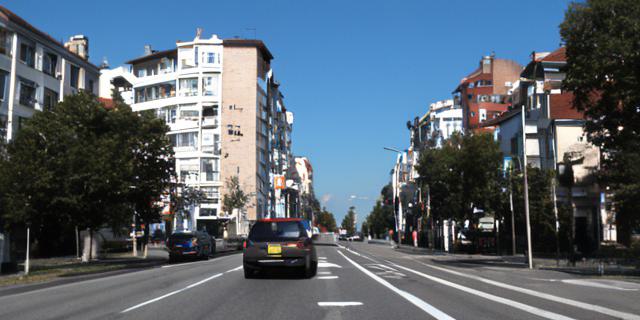}
\caption*{City}
\end{minipage}
\hfill
\begin{minipage}{.45\columnwidth}
\centering
\includegraphics[width=4cm]{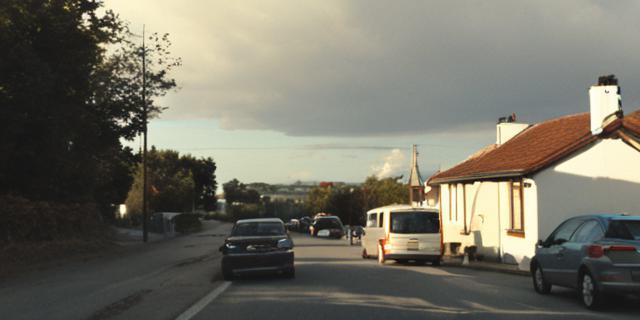}
\caption*{Rural}
\end{minipage}

\vspace{3mm}

\begin{minipage}{.45\columnwidth}
\centering
\includegraphics[width=4cm]{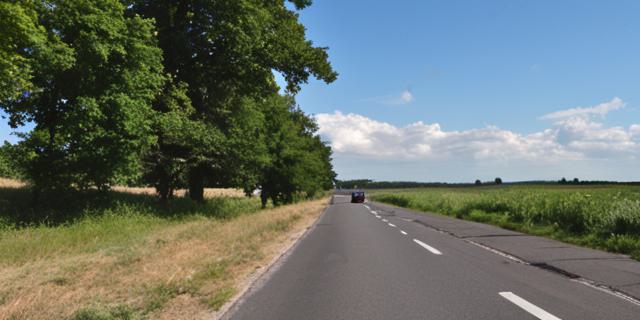}
\caption*{Field}
\end{minipage}
\hfill
\begin{minipage}{.45\columnwidth}
\centering
\includegraphics[width=4cm]{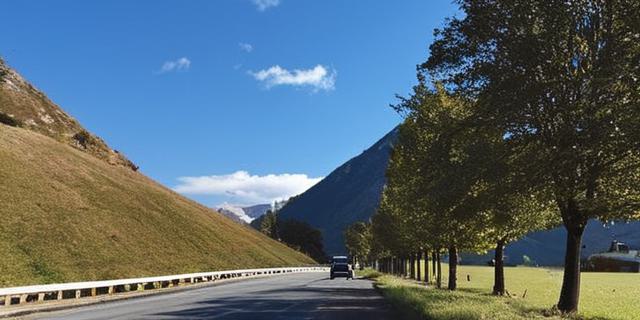}
\caption*{Mountain}
\end{minipage}
\caption{Street View Diversity}
\label{fig:street_scenarios}
\end{figure}

\begin{figure}[ht]
\centering
\begin{minipage}{.45\columnwidth}
\centering
\includegraphics[width=4cm]{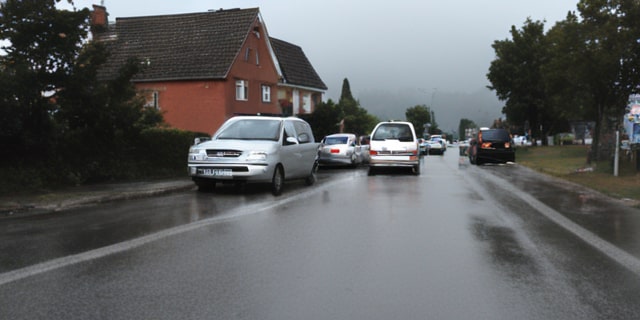}
\caption*{Heavy Traffic on a Rainy Day in Rural Areas}
\end{minipage}
\hfill
\begin{minipage}{.45\columnwidth}
\centering
\includegraphics[width=4cm]{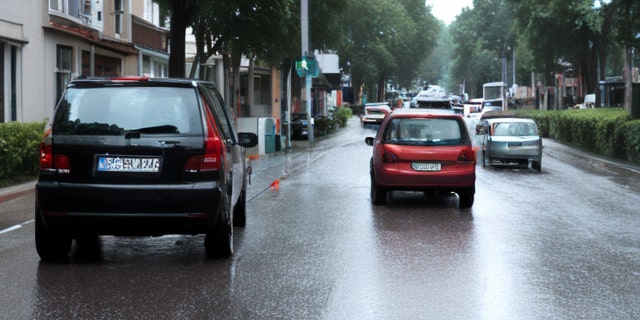}
\caption*{Heavy Traffic on a Rainy Day in City Areas}
\end{minipage}

\vspace{3mm}

\begin{minipage}{.45\columnwidth}
\centering
\includegraphics[width=4cm]{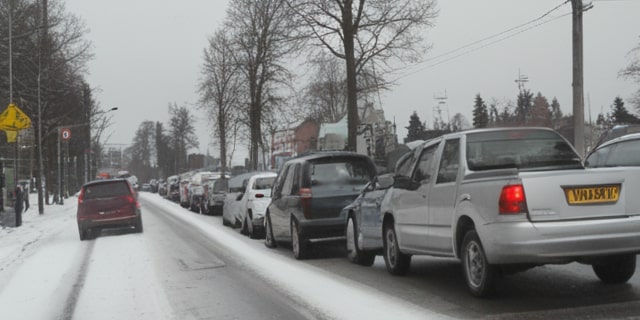}
\caption*{Heavy Traffic on a Snowy Day in Rural Areas}
\end{minipage}
\hfill
\begin{minipage}{.45\columnwidth}
\centering
\includegraphics[width=4cm]{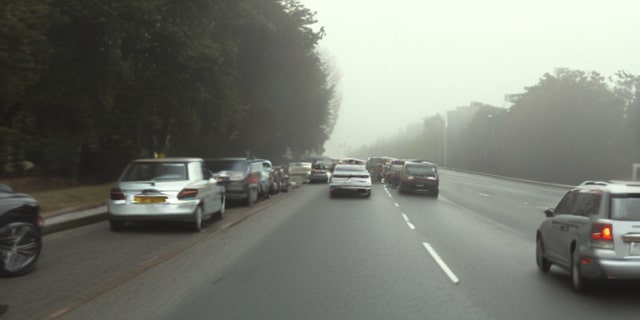}
\caption*{Heavy Traffic on a Foggy Day on the Highway}
\end{minipage}
`\caption{Generating rare driving videos}
\label{fig:rare_scenarios}
\end{figure}

While the initial series of images focused on demonstrating variability in consistent streetscapes for comparison purposes, Figure.~\ref{fig:street_scenarios} expands the scope to demonstrate the model's ability to generate different streetscapes. This illustrates the scalability and adaptability of our approach, confirming the ability of our model to simulate driving scenarios with different street layouts in addition to varying weather conditions and traffic density. The diversity of streetscapes highlights the potential of the model to generate a variety of realistic driving environments.

In Figure.~\ref{fig:rare_scenarios}, we demonstrate our model's capability to produce driving videos under rare conditions, which do not usually happen in the real world, such as ``Heavy Traffic on a Snowy Day in Rural Areas" and ``Heavy Traffic on a Foggy Day on the Highway." The videos are both realistic and high-resolution. The outcomes are promising across various conditions involving traffic density, weather, and types of streetscapes.

\section{Conclusions}\label{sec:conclusion}

This paper introduces GenDDS, a novel pipeline that leverages a stable diffusion-based model to generate driving videos under diverse conditions. Key features of GenDDS include the use of an auto-tagger mechanism for dataset preparation. Additionally, we employ LoRa to fine-tune the model and integrate ControlNet to inject additional conditions, enhancing the relevance and coherence of the generated videos. 

The evaluation results conducted on the KITTI dataset provide compelling indications that our proposed methodology is capable of generating high-resolution and realistic driving videos under different conditions or condition combinations. This work delves into the frontiers of the autonomous driving domain, focusing specifically on methods for driving video generation. With GenDDS, we can easily generate driving videos based on different conditions and settings.

To further advance the research, the generation length and complexity could be increased. Moreover, the conditions for generation could extend beyond textual descriptions of the video to include factors related to driver behavior and vehicle dynamics. Generative models can further expand the capabilities of autonomous driving technologies.

\bibliographystyle{IEEEtran}
\bibliography{ref}

\end{document}